\documentclass{article}

\usepackage[final,dblblindworkshop]{neurips_2026}
\usepackage{float}
\usepackage{placeins}

\workshoptitle{Neural Network Artifacts as a New Data Modality}

\usepackage[utf8]{inputenc}
\usepackage[T1]{fontenc}

\usepackage{hyperref}
\usepackage{url}

\usepackage{booktabs}
\usepackage{amsmath}
\usepackage{amsfonts}
\usepackage{amssymb}

\usepackage{nicefrac}
\usepackage{microtype}
\usepackage{xcolor}

\usepackage{graphicx}
\usepackage{subcaption}
\usepackage{cleveref}

\title{
Memorization in image classification:\\ test set behavior and scaling laws
}

\author{%
  Jeevesh Attri \\
  Center for Data Science \\
  New York University \\
  \texttt{ja5593@nyu.edu}
  \And
  Florentin Guth\thanks{Work done while partially affiliated with  Center for Data Science, New York University.} \\
  Center for Computational Neuroscience \\
  Flatiron Institute \\
  \texttt{fguth@flatironinstitute.org}
}

\begin{document}

\maketitle

\begin{abstract}
Deep neural networks are capable of both learning generalizing solutions on natural tasks and memorizing arbitrary mappings. We argue that understanding the former requires understanding the latter, which is much less studied. In what ways do networks that generalize or memorize differ in their behavior or their training dynamics? Do memorizing networks simply function as a lookup table where input-output mappings are stored independently? We investigate these questions in the context of CNNs trained on image classification tasks with randomized labels or pixels. We first study the behavior of such memorizing networks on unseen images. We find that independently initialized networks trained on the same random labels learn different predictive functions: they make different predictions on test examples, in contrast with generalizing networks. We then examine the effect of dataset size on memorization through the required network capacity and training time to interpolate the training set. Counter-intuitively, we find that the number of presentations of a given example required for complete memorization can \emph{decrease} with dataset size even when input-output pairs share no common structure. Although memorizing networks do not learn a universal solution, they exhibit interesting behaviors which call for further investigations into the underlying mechanisms.
\end{abstract}

\section{Introduction}
\label{sec:introduction}

A central puzzle in deep learning is to explain how neural networks generalize from training data to unseen examples. Classical statistical learning theories relate generalization to restrictions on the capacity or expressivity of the predictor \citep{vapnik1998statistical,shalev2014understanding}. Yet modern neural networks are heavily overparameterized, can improve as model size increases \citep{belkin2019reconciling,kaplan2020scaling}, and can fit arbitrary labels even when the input-label relationship contains no meaningful signal \citep{zhang2017understanding}. Parameter count and training error alone therefore do not explain why some learned solutions generalize while others merely memorize. Understanding generalization consequently requires a finer understanding of memorization: when does a network have the capacity to memorize its training set, what solution does it reach, and what are its training dynamics?

Random-label tasks provide a particularly useful setting for studying these questions. Randomizing the labels removes any meaningful relationship between inputs and targets, so that successful fitting must rely on memorization. At the same time, memorization need not reduce to independently storing arbitrary input-label pairs. Networks often learn simpler structure before fitting noise \citep{arpit2017closer}, and statistical regularities in the inputs can still influence learning under random labels \citep{maennel2020neural}. This indicates that memorizing networks may still capture non-trivial properties of the input data, and random-label training therefore provides a convenient setting in which to study memorization.

A first question is whether an analogue of generalization can be defined in pure memorization tasks through the notion of model disagreement (whether independently trained networks make similar predictions on novel inputs).  Model disagreement on unlabeled data has long been used as a signal of predictive reliability and generalization \citep{madani2004covalidation}, while more recent work has identified distributional correspondences between training and test behavior \citep{nakkiran2020distributional} and shown that the disagreement between independently trained networks indeed closely tracks their generalization error \citep{jiang2022assessing}.

Much less is known about the behavior of memorizing networks on inputs unseen during training: if independently trained networks perfectly fit the same arbitrary labels, do they also make similar predictions away from those memorized examples? For instance, this would be the case if networks predict the label of the closest training image in some (unknown) metric.

A second question concerns the behavior of memorizing networks when the size of the training set is increased. What is the capacity of a given network and how does it depend on its width? How does the time to memorize the training set depend on its size? If each input-label pair is memorized independently, the number of times a given example needs to be seen before it is memorized should not depend on the number of unrelated examples. But different examples share optimization updates (via mini-batching) and may compete for storage capacity in the network parameters, so that learning one example can slow down the learning of others. On the other hand, even if input-output relationships are independent of each other, different inputs may share a common structure (e.g., natural images have edges, textures, objects...) so that learning may benefit from additional examples. Answering these questions thus enables us to discriminate between these hypotheses and further our understanding of the underlying mechanisms of memorization in deep networks.

Our first finding is that memorizing networks reach different solutions during training: networks that perfectly memorize the same random labels agree on the training examples but remain substantially different on unseen inputs. The behavior of a memorizing network on the test set thus depends on its initialization. We also find that the number of presentations required to memorize an example \emph{decreases} with dataset size, although the image-label mapping has no shared structure across examples.
Strikingly, this effect also appears for random-pixel inputs, showing that positive interactions between memorized examples \emph{do not require shared natural-image structure}.
These results show that while memorizing networks are not universal and depend on their initialization, their training dynamics exhibit surprising phenomena that call for further explanations.

Code and scripts for reproducing all experiments and figures are available at \url{https://github.com/BabylonBhardwaj/NeurIPS_Workshop}.

\section{Test-set behavior across learning and memorization conditions}
\label{sec:results_predictive}

We consider convolutional networks trained on CIFAR-10 \citep{krizhevsky2009learning} with cross-entropy loss and SGD. When scaling the width of our networks, we use the so-called maximal-update parameterization ($\mu$P) \citep{greg-yang-tensor-programs-iv,simon-bernstein-yang}. We study three data conditions designed to separate the effects of structure in the target function or the input data distribution. In the \textit{true-label} condition, we use the original CIFAR-10 images and labels, which serves as a baseline against the following two memorization tasks. In the \textit{random-label} condition, the natural images are retained while the training labels are replaced by a fixed random assignment sampled uniformly from the ten classes. In the \textit{random-pixel} condition, the images are replaced by independently sampled pixel arrays (labels are then again statistically independent from input images). We provide more experimental details in Appendix~\ref{app:experimental_details}.

To study whether independently trained networks converge to similar predictive behavior, we train ensembles of 20 independently initialized CNNs (comprising three convolutional layers with 128/256/512 channels and a 1024-unit hidden layer) on the full CIFAR-10 training set under each of the three data conditions: true labels, random labels, and random pixels. All models are trained with a batch size of 128 and a base learning rate of 0.01 for at most 100 epochs or after nearly interpolating the training set (more details in Appendix~\ref{app:experimental_details}). After every epoch, we record the full predictive distribution of each model on both the training and test sets, of which we report several summary statistics below.

We first ask whether independently initialized networks trained under the three data conditions make similar predictions on unseen inputs. Classification networks output a probability distribution over labels for each input. To probe properties of these distributions, we track three quantities across both train and test sets: label fit, measured by cross-entropy with the ground truth (negative log probability of the correct label); uncertainty, as measured by the entropy of the predicted label distribution; and disagreement between models, as measured by the Jensen--Shannon divergence between their predicted label distributions.

\begin{figure}[!ht]
    \centering
    \includegraphics[width=\linewidth]{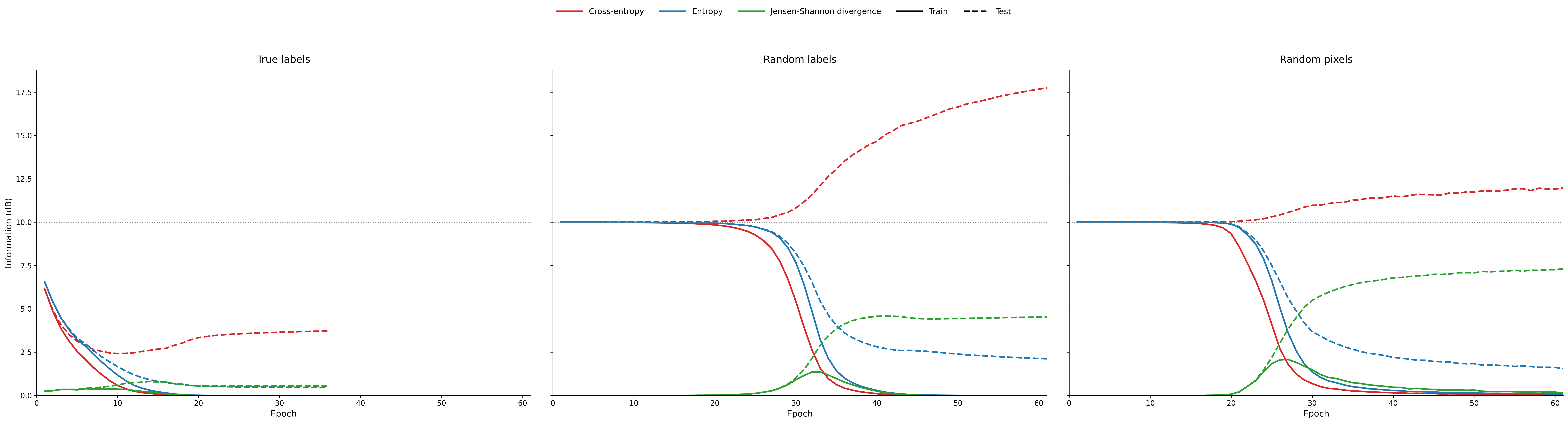}
    \caption{
    Evolution of label fit, uncertainty, and disagreement across true-label (left), random-label (center), and random-pixel (right) conditions. Solid lines show training-set behavior and dashed lines test-set behavior. Cross-entropy measures label fit (red), entropy measures predictive uncertainty (blue), and Jensen--Shannon divergence measures disagreement across independently initialized models (green), defined as $\mathrm{JSD}=H(\bar p)-\frac{1}{M}\sum_{m=1}^{M}H(p_m)$, where $\bar p=\frac{1}{M}\sum_{m=1}^{M}p_m$ (see Appendix~\ref{app:exp1}). Quantities are reported in dB, with $10$dB corresponding to uniform prediction (dotted gray line); JSD is rescaled to have a maximum value of $10$dB. Results are averaged over 20 independently initialized $4\times$-width models per condition (see Appendix~\ref{app:architecture}).
}
    \label{fig:predictive_dynamics}
\end{figure}

Figure~\ref{fig:predictive_dynamics} shows a clear contrast between learning and memorization. With true labels, disagreement remains low on both training and test examples, despite some divergence between their predictive uncertainties and label fit. Under random labels, the models temporarily diverge while memorizing before converging almost perfectly on the training examples; on unseen inputs, however, they become increasingly confident while disagreeing more strongly with one another, and test accuracy remains near chance. Random-pixel inputs show an even stronger separation: training predictions converge while test-set disagreement rises substantially as individual predictions become more confident. For random labels, additional experiments (not shown) confirmed that the predicted label on test examples is uniformly distributed when varying the random initialization. Thus, convergence on memorized training examples does not imply convergence to a common predictive function away from them, and this effect persists even in the absence of natural-image structure. We present an alternative visualization of these predictive dynamics in Appendix~\ref{app:entropy_jsd}, Figure~\ref{fig:entropy_jsd}.

\FloatBarrier

\section{Memorization capacity frontiers}
\label{sec:results_scaling}

We next examine how memorization depends on model width, dataset size, input structure, and optimization policy.
We find the strongest dataset-size dependence appearing for random-label and random-pixel inputs.

To study how memorization efficiency changes with dataset size and optimization procedure, we train models on nested fractions of CIFAR-10 of size $2^{-4}$ to $2^0$ (full dataset).
We evaluate width multipliers $m \in \{0.0625,\ 0.125,\ 0.25,\ 0.5,\ 1,\ 2,\ 4,\ 8\}$. A run terminates when evaluated training accuracy first reaches 100\% or after 100 epochs.

We compare two joint batch-size and learning-rate policies. Under the \textit{fixed} policy, batch size and base learning rate remain constant across dataset fractions, so larger datasets lead to more sequential parameter updates per epoch. Under the \textit{fraction-scaled} policy, both instead increase with dataset size, keeping updates per epoch approximately constant. Comparing the two allows testing separately the effects of batch size and number of gradient steps on memorization efficiency.

Remaining implementation details are given in Appendix~\ref{app:experimental_details}.

\begin{figure}[!ht]
    \centering
    \includegraphics[width=\linewidth]{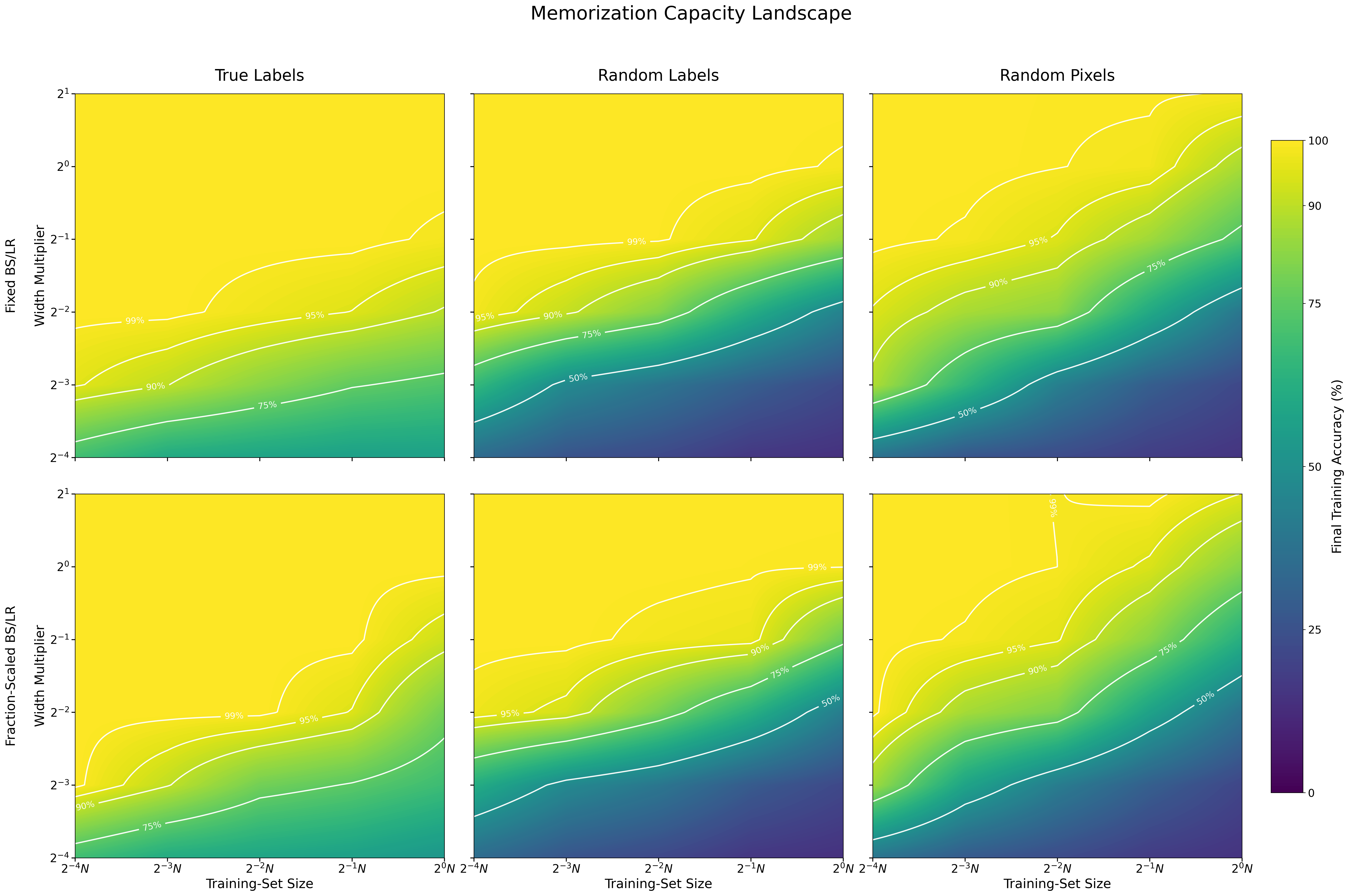}
    \caption{
        Memorization capacity frontiers across model width and training-set fraction.
        Rows correspond respectively to fixed and fraction-scaled batch-size/learning-rate policies,
        and columns to true labels, random labels, and random-pixel inputs.
        The final training accuracy in the width-training set size plane is estimated by interpolation from its value on grid points.
        }
    \label{fig:capacity_frontier}
    \vspace{-0.5em}
\end{figure}

Figure~\ref{fig:capacity_frontier} shows that true labels can be fitted across a broad range of widths and dataset sizes, while random labels require progressively greater capacity as the training set grows. Random-pixel inputs are the most demanding, especially under the fraction-scaled policy, where incomplete memorization persists even at comparatively large widths. The memorization frontiers are approximately linear in log-width versus log-dataset-size. Under the fixed policy, slopes are roughly 0.2-0.4 for true labels, 0.4 for random labels, and 0.5-0.7 for random pixels. Under the fraction-scaled policy, slopes are roughly 0.3-0.6 for true labels, 0.35-0.48 for random labels, and 0.54-0.93 for random pixels, with the steepest dependence appearing for random pixels.

\FloatBarrier

\section{Memorization efficiency}
\label{sec:memorization_efficiency}

The main comparisons use the same architecture as Section~\ref{sec:results_predictive}, with additional width experiments reported in Appendix~\ref{app:experimental_details}. Since each epoch presents every selected example once, the epoch at which evaluated training accuracy reaches 100\% is also the number of presentations per example, our primary measure of memorization efficiency.

\begin{figure}[!ht]
\centering
\includegraphics[width=\linewidth]{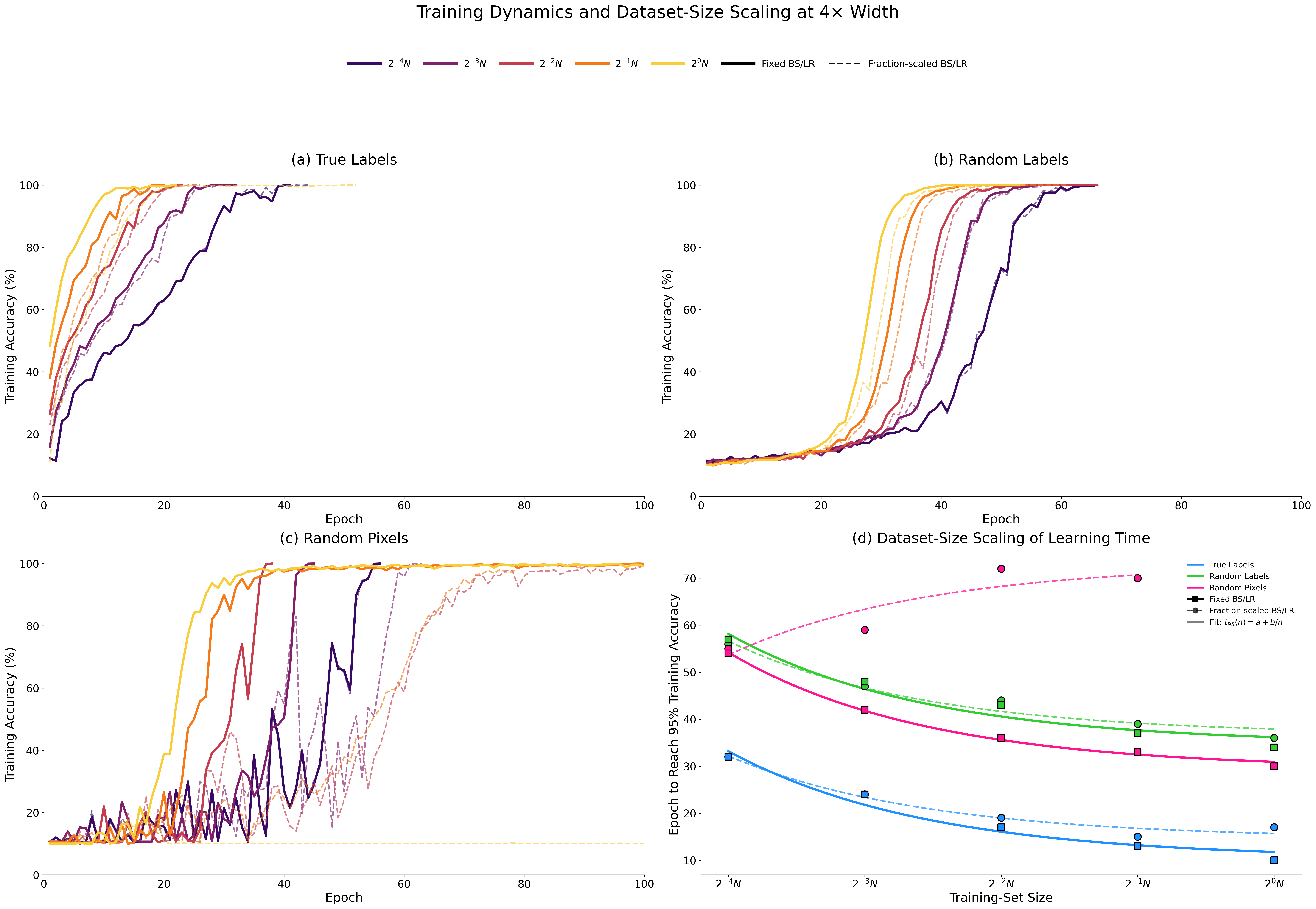}
\caption{
Training dynamics across dataset sizes, memorization tasks, and training strategies for a 4$\times$-width model. Panels (a-c) show evaluated training accuracy across epochs for true labels, random labels, and random-pixel inputs at five training-set sizes (indicated by line color). Solid lines denote the fixed batch-size/learning-rate policy, and dashed lines denote the fraction-scaled policy. Panel (d) shows the epoch at which training accuracy first reaches $95\%$ as a function of training-set size, together with fitted curves $t_{95}(n) = a + b/n$.
}
\label{fig:memorization_efficiency}
\end{figure}

We now report the training time needed for the 4$\times$ width model to memorize the training set as a function of its size in Figure~\ref{fig:memorization_efficiency}. Because every epoch presents each selected example once, epoch count is also the number of presentations per example. True-label training is consistently fastest while random labels are learned more slowly, but all five fractions are memorized by the 4$\times$ model under both policies. In both cases, the number of required presentations per example is not constant but \emph{decreases} with dataset size. This is expected for true labels, as each additional example is informative about the general task (learning could be done in a single epoch on a large dataset), but is much more surprising for random labels which do not share this property. This indicates that even arbitrary associations are not learned independently of the surrounding dataset. This effect is observed across both training policies, with slightly lower epoch counts for fixed (rather than scaled) batch size and learning rate.

The strongest policy dependence appears for random-pixel inputs. Under the fixed policy, the presentations required for memorization fall substantially as the dataset grows. Under the fraction-scaled policy, this advantage disappears at larger fractions, and the $2^{-2}$- to $2^0$-fraction models do not reach complete memorization within the training budget. Several random-pixel trajectories are also irregular, and we have observed that training can become unstable in this case.

To quantify the memorization efficiency, we extract from each training curve the number of epochs to reach $95\%$ training accuracy, and observe that it is well-modeled by an expression of the form $t_{95}(n) = a + b/n$. In other words, the number of presentations to memorize an individual example decays inversely with dataset size towards a limit value corresponding to infinitely large datasets $t_{95}(\infty) = a$.

Strikingly, under the fixed policy, the $b$ values are almost constant across true labels, random labels, and random pixels (they all lie in the range $7.2$--$7.8\times10^4$), even though the limiting values $a$ differ substantially across tasks: approximately $10.3$, $34.7$, and $29.4$ presentations, respectively. The much smaller limiting value for true labels is consistent with ordinary generalization, where sufficiently large datasets reduce the need to repeatedly present any particular example.

Related work has also studied other aspects of memorization. Instance-specific analyses have related interpolation difficulty to the geometry of the dataset \citep{dirksen2024memorization}, while other work has linked memorization to local loss curvature and to conflicting or mislabeled examples \citep{garg2024curvature}. Recent results for graph neural networks likewise show that the amount of memorization changes with how informative the underlying input structure is \citep{jamadandi2025memorization}.

\FloatBarrier

\section{Discussion}
\label{sec:discussion_limitations}

In this work, we have explored fundamental properties of memorizing neural networks on randomized tasks and discovered several surprising phenomena. First, unlike networks trained on true labels, networks memorizing arbitrary mappings do not converge to a common predictive function on unseen inputs: their test predictions become high-confidence but remain strongly initialization-dependent. This seems to indicate that neural networks do not exploit any inductive bias when they memorize arbitrary mappings, in stark contrast with standard learning where generalization occurs. However, these arbitrary input-label pairs are not memorized completely independently: adding unrelated gradient updates can reduce the number of presentations required to memorize an association, while increasing the batch size can weaken or even reverse this benefit. The fact that this interaction persists for random-pixel inputs is particularly informative, because it shows that the effect does not require shared semantic or natural-image structure. These results show that memorizing networks exhibit interesting behavior that is markedly different from the more commonly-studied phenomenon of generalization, and call for further explanation.

These findings raise several questions for future work. Most directly, what properties of the training dynamics produce the positive interaction between otherwise arbitrary examples? The present experiments establish that the effect exists independently of the data distribution and that it changes with batch size scaling, but they do not identify its mechanism. It would therefore be useful to test whether the interaction can be characterized through gradient alignment, properties of the optimization trajectory, or other measurable relationships between examples. More broadly, future work should determine whether similar or different phenomena appear across architectures, optimizers, data modalities, and larger models, and whether they extend from deliberately randomized tasks to naturally occurring forms of memorization (e.g., facts in LLMs).

\clearpage
\bibliographystyle{plainnat}
\bibliography{references}

@book{vapnik1998statistical,
  title     = {Statistical Learning Theory},
  author    = {Vapnik, Vladimir N.},
  publisher = {Wiley},
  year      = {1998}
}

@book{shalev2014understanding,
  title     = {Understanding Machine Learning: From Theory to Algorithms},
  author    = {Shalev-Shwartz, Shai and Ben-David, Shai},
  publisher = {Cambridge University Press},
  year      = {2014}
}

@article{belkin2019reconciling,
  title   = {Reconciling Modern Machine-Learning Practice and the Classical Bias--Variance Trade-Off},
  author  = {Belkin, Mikhail and Hsu, Daniel and Ma, Siyuan and Mandal, Soumik},
  journal = {Proceedings of the National Academy of Sciences},
  volume  = {116},
  number  = {32},
  pages   = {15849--15854},
  year    = {2019}
}

@article{kaplan2020scaling,
  title   = {Scaling Laws for Neural Language Models},
  author  = {Kaplan, Jared and McCandlish, Sam and Henighan, Tom and Brown, Tom B. and Chess, Benjamin and Child, Rewon and Gray, Scott and Radford, Alec and Wu, Jeffrey and Amodei, Dario},
  journal = {arXiv preprint arXiv:2001.08361},
  year    = {2020}
}

@inproceedings{zhang2017understanding,
  title     = {Understanding Deep Learning Requires Rethinking Generalization},
  author    = {Zhang, Chiyuan and Bengio, Samy and Hardt, Moritz and Recht, Benjamin and Vinyals, Oriol},
  booktitle = {International Conference on Learning Representations},
  year      = {2017}
}

@inproceedings{arpit2017closer,
  title     = {A Closer Look at Memorization in Deep Networks},
  author    = {Arpit, Devansh and Jastrz{\k{e}}bski, Stanis{\l}aw and Ballas, Nicolas and Krueger, David and Bengio, Emmanuel and Kanwal, Maxinder S. and Maharaj, Tegan and Fischer, Asja and Courville, Aaron and Bengio, Yoshua and Lacoste-Julien, Simon},
  booktitle = {Proceedings of the 34th International Conference on Machine Learning},
  series    = {Proceedings of Machine Learning Research},
  volume    = {70},
  pages     = {233--242},
  year      = {2017},
  publisher = {PMLR}
}

@inproceedings{maennel2020neural,
  title     = {What Do Neural Networks Learn When Trained With Random Labels?},
  author    = {Maennel, Hartmut and Alabdulmohsin, Ibrahim M. and Tolstikhin, Ilya O. and Baldock, Robert and Bousquet, Olivier and Gelly, Sylvain and Keysers, Daniel},
  booktitle = {Advances in Neural Information Processing Systems},
  volume    = {33},
  year      = {2020}
}

@inproceedings{jiang2022assessing,
  title     = {Assessing Generalization of SGD via Disagreement},
  author    = {Jiang, Yiding and Nagarajan, Vaishnavh and Baek, Christina and Kolter, J. Zico},
  booktitle = {International Conference on Learning Representations},
  year      = {2022}
}

@inproceedings{greg-yang-tensor-programs-iv,
  title     = {Tensor Programs IV: Feature Learning in Infinite-Width Neural Networks},
  author    = {Yang, Greg and Hu, Edward J.},
  booktitle = {Proceedings of the 38th International Conference on Machine Learning},
  series    = {Proceedings of Machine Learning Research},
  volume    = {139},
  pages     = {11727--11737},
  year      = {2021},
  publisher = {PMLR},
  url       = {https://proceedings.mlr.press/v139/yang21c.html}
}

@inproceedings{simon-bernstein-yang,
  title     = {Tensor Programs V: Tuning Large Neural Networks via Zero-Shot Hyperparameter Transfer},
  author    = {Yang, Greg and Hu, Edward J. and Babuschkin, Igor and Sidor, Szymon and Liu, Xiaodong and Farhi, David and Ryder, Nick and Pachocki, Jakub and Chen, Weizhu and Gao, Jianfeng},
  booktitle = {Advances in Neural Information Processing Systems},
  volume    = {34},
  year      = {2021}
}

@inproceedings{madani2004covalidation,
  title     = {Co-Validation: Using Model Disagreement on Unlabeled Data to Validate Classification Algorithms},
  author    = {Madani, Omid and Pennock, David M. and Flake, Gary W.},
  booktitle = {Advances in Neural Information Processing Systems},
  volume    = {17},
  year      = {2004}
}

@article{nakkiran2020distributional,
  title   = {Distributional Generalization: A New Kind of Generalization},
  author  = {Nakkiran, Preetum and Bansal, Yamini},
  journal = {arXiv preprint arXiv:2009.08092},
  year    = {2020}
}

@techreport{krizhevsky2009learning,
  title       = {Learning Multiple Layers of Features from Tiny Images},
  author      = {Krizhevsky, Alex},
  institution = {University of Toronto},
  year        = {2009},
  url         = {https://www.cs.toronto.edu/~kriz/learning-features-2009-TR.pdf}
}

@article{dirksen2024memorization,
  title   = {Memorization With Neural Nets: Going Beyond the Worst Case},
  author  = {Dirksen, Sjoerd and Finke, Patrick and Genzel, Martin},
  journal = {Journal of Machine Learning Research},
  volume  = {25},
  number  = {347},
  pages   = {1--38},
  year    = {2024}
}

@inproceedings{garg2024curvature,
  title     = {Memorization Through the Lens of Curvature of Loss Function Around Samples},
  author    = {Garg, Isha and Ravikumar, Deepak and Roy, Kaushik},
  booktitle = {Proceedings of the 41st International Conference on Machine Learning},
  series    = {Proceedings of Machine Learning Research},
  volume    = {235},
  pages     = {15083--15101},
  year      = {2024},
  publisher = {PMLR}
}

@inproceedings{jamadandi2025memorization,
  title     = {Memorization in Graph Neural Networks},
  author    = {Jamadandi, Adarsh and Xu, Jing and Dziedzic, Adam and Boenisch, Franziska},
  booktitle = {Advances in Neural Information Processing Systems},
  volume    = {38},
  year      = {2025}
}

\appendix

\section{Experimental details}
\label{app:experimental_details}

\subsection{Compute resources}

All reported experiments were run on Google Colab using a single
NVIDIA A100-SXM4-80GB GPU with 80 GB of GPU memory. The runtime
environment provided approximately 167 GiB of system memory and
23 GB of local storage.

The memorization-frontier experiment comprised 240 individual
training runs. Recorded execution times ranged from approximately
15.8 seconds to 17.4 minutes per run, with a median of 63.5 seconds
and a mean of 171.0 seconds. Across all 240 runs, the recorded
execution time was approximately 11.4 GPU-hours.

The predictive-dynamics experiment comprised three ensembles of
20 independently initialized models, one for each data condition.
The true-label, random-label, and random-pixel experiments required
approximately 52.2, 90.8, and 98.1 minutes, respectively, for a total
of approximately 4.02 GPU-hours. The experiments reported in this
paper therefore required approximately 15.4 A100 GPU-hours in total.

Additional preliminary and exploratory experiments were conducted
during development but are not included in this compute estimate.

\subsection{Architecture and optimization}
\label{app:architecture}

We use the standard CIFAR-10 dataset \citep{krizhevsky2009learning}, obtained from the original University of Toronto distribution and used in accordance with the terms provided by the dataset creators.

Both experiments use a width-scaled convolutional network. For width multiplier $m$, the three convolutional layers contain $32m$, $64m$, and $128m$ channels, respectively. Each convolutional layer is followed by a ReLU activation and $2\times2$ max pooling. The resulting representation is passed through a fully connected hidden layer with $256m$ units and a ten-class output layer.

We use maximal-update parameterization ($\mu$P) with base width 1 and delta width 2. Models are trained with cross-entropy loss using $\mu$P-scaled SGD with momentum 0.9. CIFAR-10 images are converted to tensors and normalized channel-wise with mean 0.5 and standard deviation 0.5. No data augmentation is used.

Unless otherwise specified, training and evaluation use the standard CIFAR-10 training and test sets. Random-label and random-pixel transformations are applied only in their corresponding experimental conditions.

\subsection{Predictive-dynamics experiment}
\label{app:exp1}

We train three ensembles of 20 independently initialized $4\times$-width networks, one under each of the true-label, random-label, and random-pixel conditions, on the full 50,000-example CIFAR-10 training set. Each ensemble uses model initialization seeds 1000--1019. In the random-label condition, labels are sampled uniformly from the ten classes using label seed 42. In the random-pixel condition, CIFAR-10 images are replaced by independently sampled pixel arrays while retaining the original class labels. Within each condition, all models use identical data and optimization settings and differ only in initialization. Training uses batch size 128, base learning rate 0.01, and momentum 0.9, with evaluation on the complete training and corresponding test sets after every epoch.

Training proceeds for at most 100 epochs. Early stopping requires essentially complete training-set fit and predictive agreement on the training set: mean training loss and its across-model standard deviation must each be at most $10^{-4}$; mean training accuracy must be at least 99.999\% with across-model standard deviation at most $10^{-3}$; ensemble training accuracy must be at least 99.999\%; ensemble entropy and mean individual entropy must each be at most 0.01 dB; Jensen-Shannon divergence must be at most 0.001 dB; and ensemble label negative log-likelihood must be at most 0.01 dB.

For each example, let $p_m$ denote the predictive distribution of model $m$. The ensemble-mean distribution is

\begin{equation}
\bar{p} = \frac{1}{M}\sum_{m=1}^{M} p_m.
\end{equation}

We compute ensemble entropy $H(\bar{p})$, mean individual entropy

\begin{equation}
\frac{1}{M}\sum_{m=1}^{M} H(p_m),
\end{equation}

and Jensen--Shannon divergence

\begin{equation}
\operatorname{JSD} = H(\bar{p}) - \frac{1}{M}\sum_{m=1}^{M} H(p_m).
\end{equation}

We additionally record ensemble label negative log-likelihood, mean pairwise symmetric KL divergence, model-to-ensemble KL divergence, individual and ensemble accuracy, predictive confidence, and modal class agreement.

Information quantities are reported in decibels using base-10 logarithms with a factor of 10. Under this convention, a uniform predictive distribution over the ten CIFAR-10 classes has entropy 10 dB. Test-set label negative log-likelihood is evaluated against the original CIFAR-10 test labels.

\subsection{Memorization-frontier experiment}
\label{app:exp2}

The memorization-frontier experiment contains three data conditions, five training-set fractions, eight model widths, and two joint batch-size/learning-rate policies, for a total of

\begin{equation}
3 \times 5 \times 8 \times 2 = 240
\end{equation}

training runs.

The training-set fractions are

\begin{equation}
f \in \left\{ 0.0625,\ 0.125,\ 0.25,\ 0.5,\ 1.0 \right\},
\end{equation}

corresponding to 3,125, 6,250, 12,500, 25,000, and 50,000 CIFAR-10 training examples. Subsets are nested: a single permutation of the full training set is generated using subset seed 999, and each fraction uses the corresponding prefix. Examples appearing in a smaller fraction therefore also appear in every larger fraction.

We evaluate width multipliers

\begin{equation}
m \in \left\{ 0.0625,\ 0.125,\ 0.25,\ 0.5,\ 1,\ 2,\ 4,\ 8 \right\}.
\end{equation}

The three data conditions are constructed as follows. \textbf{True labels} use the original CIFAR-10 images and labels. \textbf{Random labels} retain the natural CIFAR-10 images but replace the training labels with a fixed random assignment sampled uniformly over the ten classes. Because the dataset fractions are nested, overlapping examples retain the same random label across fractions. \textbf{Random pixels} replace the CIFAR-10 images with independently sampled pixel arrays while retaining the original class labels. Independent random-pixel arrays are generated for the training and test sets.

We compare two joint batch-size and base-learning-rate policies. Under the \textbf{fixed policy}, batch size is 108 and base learning rate is 0.01 at every training fraction. The resulting numbers of minibatch updates per epoch are approximately 29, 58, 116, 232, and 463 for fractions 0.0625, 0.125, 0.25, 0.5, and 1.0, respectively.

Under the \textbf{fraction-scaled policy}, batch size and base learning rate increase proportionally with training fraction:

\begin{center}
\begin{tabular}{ccc}
\toprule
Training fraction & Batch size & Base learning rate \\
\midrule
0.0625 & 108  & 0.01 \\
0.125  & 216  & 0.02 \\
0.25   & 432  & 0.04 \\
0.5    & 864  & 0.08 \\
1.0    & 1728 & 0.16 \\
\bottomrule
\end{tabular}
\end{center}

These settings keep the number of minibatch updates per epoch approximately constant across fractions, up to the final partial minibatch.

For a given data condition, fraction, and width, the fixed and fraction-scaled runs use the same model-initialization seed. Initialization is therefore matched across the two policies for directly corresponding configurations; initialization seeds are not held fixed across different dataset fractions.

After every training epoch, we evaluate accuracy on the complete selected training subset and on the corresponding test condition. A run terminates when evaluated training accuracy first reaches 100\% or after 100 epochs. Runs that do not reach complete memorization within this budget are treated as not memorized within the observed training horizon rather than as having a memorization time of exactly 100 epochs.

Because every epoch presents every selected example once, epoch count is also the number of presentations per example. Figure~\ref{fig:capacity_frontier} uses final training accuracy to visualize the width-by-fraction capacity landscape, while Figure~\ref{fig:memorization_efficiency} shows the detailed 4$\times$ training trajectories across the five dataset fractions.

\subsection{Dataset-size learning-time summary}
\label{app:dataset_scaling}

To summarize the dataset-size dependence of the 4$\times$ runs without relying on exact saturation at 100\% accuracy, we record the first epoch at which evaluated training accuracy reaches 95\%,

\begin{equation}
t_{95} = \min\{\, e : a(e) \geq 95\% \,\},
\end{equation}

where $a(e)$ denotes evaluated training accuracy at epoch $e$. Runs that do not reach 95\% within the 100-epoch budget are treated as not reaching the threshold within the observed horizon rather than being assigned a value of 100.

For each data condition and optimization policy, the observed learning times are summarized with the descriptive relation

\begin{equation}
t_{95}(n) = a + \frac{b}{n},
\end{equation}

where $n$ is the number of training examples. The fit, shown in Figure~\ref{fig:memorization_efficiency}(d), is intended only as a compact description of the measured dataset-size trend and is not interpreted as a universal or asymptotic scaling law.

The random-pixel trajectories are notably less regular than those for true and random labels, and several runs fail to reach the 95\% threshold within the training budget. Since each configuration is represented by a single run, these irregularities may partly reflect stochastic variation. We therefore use the fitted curves to summarize broad differences across dataset sizes and optimization policies rather than to support precise extrapolation beyond the observed settings.

\clearpage
\section{Alternative visualization of predictive dynamics}
\label{app:entropy_jsd}

\begin{figure}[!ht]
    \centering
    \includegraphics[width=\linewidth]{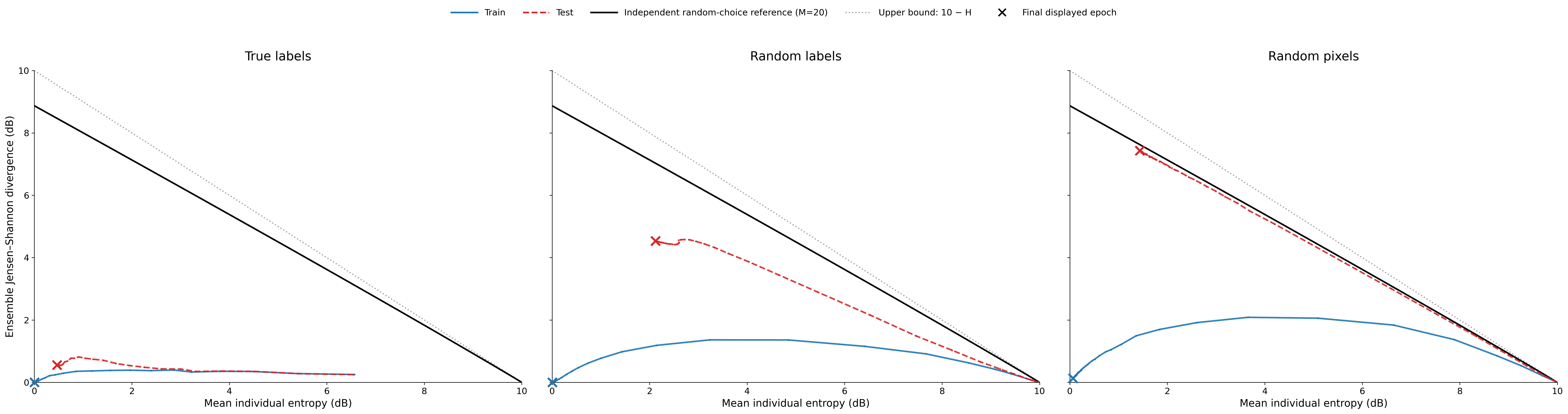}
    \caption{
    Ensemble Jensen--Shannon divergence versus mean individual
    predictive entropy for true labels, random labels, and random
    pixels. Blue solid curves show training trajectories and red
    dashed curves show test trajectories, with points connected
    in epoch order. Crosses mark the final recorded epoch.
    The black curve gives the expected ensemble JSD for 20 models
    with equal confidence and independently chosen uniform
    preferred classes. The dotted gray line shows the upper
    bound $10-H$, where $H$ is mean individual entropy in dB.
    The reference is descriptive and does not constitute a test
    of independence or uniformity.
    }
    \label{fig:entropy_jsd}
\end{figure}

Figure~\ref{fig:entropy_jsd} presents an alternative visualization
of Figure~\ref{fig:predictive_dynamics}, plotting ensemble
Jensen--Shannon divergence against mean individual predictive
entropy. Training trajectories approach low entropy and low
disagreement in all three conditions. On test inputs,
random-pixel trajectories closely follow the independent
random-choice reference, while random-label trajectories
remain below it and true-label trajectories show substantially
less disagreement.

\end{document}